\documentclass[letterpaper, 10 pt, journal, twoside]{IEEEtran} 

\usepackage{times}

\usepackage{multicol}

\usepackage{moreverb,url}

\usepackage[colorlinks,bookmarksopen,bookmarksnumbered,citecolor=red,urlcolor=red]{hyperref}

\usepackage{hyperref}
\usepackage[nolist,nohyperlinks]{acronym}
\usepackage{amsmath,amsfonts,amssymb}
\usepackage{tikz,tkz-euclide}
\usetikzlibrary{arrows,calc,shapes,positioning}
\tikzset{>=latex}
\usepackage{color,soul}
\usepackage{graphicx}
\usepackage{calc}
\usepackage{pgfplots}
\usepackage{colortbl}
\usepackage{todonotes}
\usepackage{multirow}
\usepackage{mathtools}
\def\identitymat{\mathbf{I}}

\newcommand{\units}[1]{\,\mathrm{#1}}
\def\reals{\mathbb{R}}
\def\surface{\mathcal{S}}
\def\occ{o}
\def\distfield{{\hat{d}}}

\def\dist{d}
\def\revert{r}

\def\matern{{Mat\'{e}rn} }
\def\lengthscale{l}
\def\gp{\mathcal{GP}}
\def\gaussian{\mathcal{N}}

\def\ordinate{y}
\def\ordinatevec{\mathbf{y}}

\def\abscissavec{\mathbf{x}}
\def\abscissamat{\mathbf{X}}

\newcommand{\kernelraw}[1]{{{k}_{#1}}}
\newcommand{\kernel}[3]{{{k}_{#1}\left(#2,#3\right)}}

\newcommand{\kernelmatraw}[1]{{\mathbf{K}_{#1}}}

\newcommand{\kernelvecraw}[1]{{\mathbf{k}_{#1}}}

\def\gradient{\nabla}

\newcommand\pos[2]{{\mathbf{p}_{#1}^{#2}}}

\def\noise{{\eta}}

\usepackage{tabularx}

\title{Accurate Gaussian-Process-based Distance Fields with applications to Echolocation and Mapping}

\author{Cedric Le Gentil$^{1}$,
Othmane-Latif Ouabi$^{2}{}^{3}$,
Lan Wu$^{1}$, 
Cedric Pradalier$^{3}$ and
Teresa Vidal-Calleja$^{1}$%
\thanks{Cedric Le Gentil and Teresa Vidal-Calleja are supported by the Australian Research Council Discovery Project under Grant DP210101336.
This work has been partially funded by the BugWright2 project, supported by the European Commission under grant agreement 871260 - BugWright2.}
\thanks{$^{1}$Robotics Institute at the University of Technology Sydney,
Australia {\tt\small \{cedric.legentil; lan.wu-2; teresa.vidalcalleja \}@uts.edu.au}}%
\thanks{$^{2}$Sysnav, France {\tt\small othmane.ouabi@sysnav.fr}}%
\thanks{$^{3}$International Research Lab 2958 Georgia Tech-CNRS, France {\tt\small cedricp@georgiatech-metz.fr}}%
\thanks{© 2023 IEEE.  Personal use of this material is permitted.  Permission from IEEE must be obtained for all other uses, in any current or future media, including reprinting/republishing this material for advertising or promotional purposes, creating new collective works, for resale or redistribution to servers or lists, or reuse of any copyrighted component of this work in other works.}
}

\begin{document}

\maketitle
\thispagestyle{empty}
\pagestyle{empty}

\begin{abstract}
This paper introduces a novel method to estimate distance fields from noisy point clouds using Gaussian Process (GP) regression.
Distance fields, or distance functions, gained popularity for applications like point cloud registration, odometry, SLAM, path planning, shape reconstruction, etc.
A distance field provides a continuous representation of the scene defined as the shortest distance from any query point and the closest surface.
The key concept of the proposed method is the transformation of a GP-inferred latent scalar field into an accurate distance field by using a \emph{reverting} function related to the kernel inverse.
The latent field can be interpreted as a smooth occupancy map.
This paper provides the theoretical derivation of the proposed method as well as a novel uncertainty proxy for the distance estimates.
The improved performance compared with existing distance fields is demonstrated with simulated experiments.
The level of accuracy of the proposed approach enables novel applications that rely on precise distance estimation: this work presents echolocation and mapping frameworks for ultrasonic-guided wave sensing in metallic structures.
These methods leverage the proposed distance field with a physics-based measurement model accounting for the propagation of the ultrasonic waves in the material.
Real-world experiments are conducted to demonstrate the soundness of these frameworks.
\end{abstract}


\section{Introduction}

Robotic perception and map representations are the key components of any autonomous system operating in the real world.
There exist many representations to model the environment in which a system is running, each with its own specific set of applications.
For example, sparse geometric features are traditionally used for \ac{vo} and \ac{slam} \cite{Cadena2016,paper:orbslam2}, occupancy grids \cite{occupancy} allow for dense mapping and path planning \cite{Octomap, occupancy2016review}, etc.
Unfortunately, the multitude of representations needed to run a single robot in the real world leads to a high level of complexity and hinders holistic approaches.
In this paper, we introduce an accurate continuous distance field representation based on \ac{gp} regression that accommodates a wide variety of applications such as localisation, mapping, and planning within a unique representation.

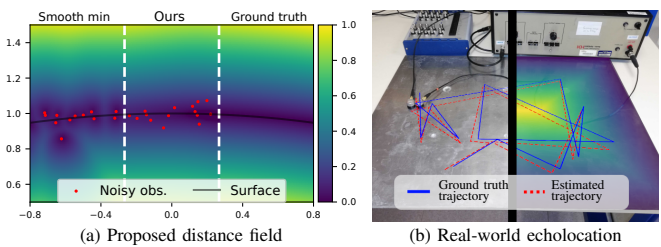
\begin{figure}
    \centering
    \begin{tikzpicture}
        \node[inner sep=0,outer sep=0] (bat) {\includegraphics[clip, trim=0.35cm 0.6cm 0.35cm 0.6cm, height=2.8cm]{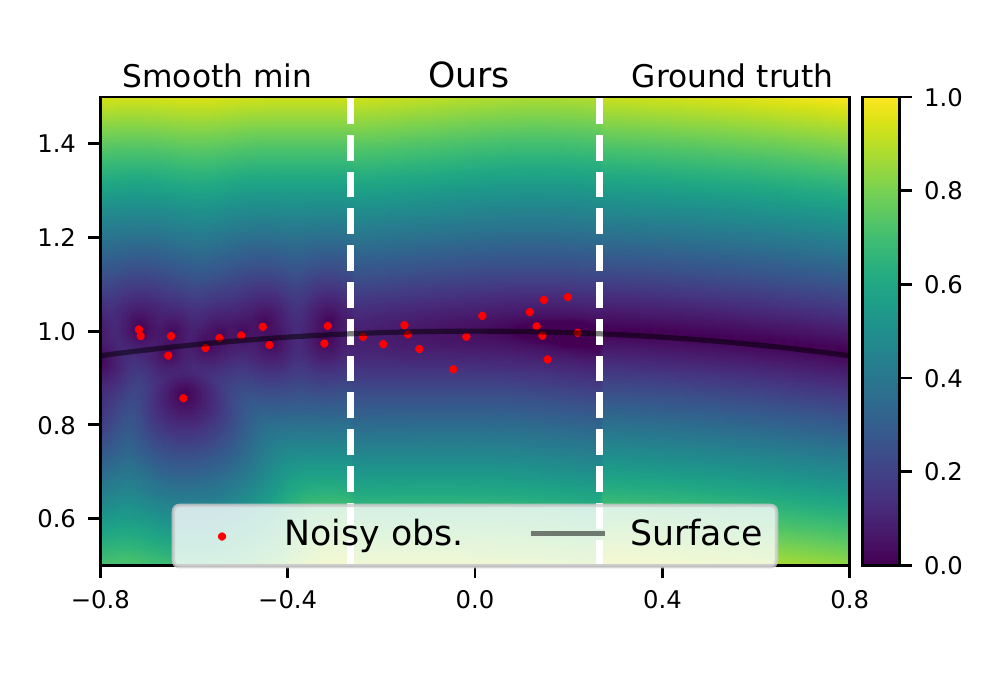}};
        \node[right=0.2cm of bat,inner sep=0,outer sep=0] (real) {\includegraphics[clip, trim=1.5cm 0cm 1cm 2cm, height=2.8cm]{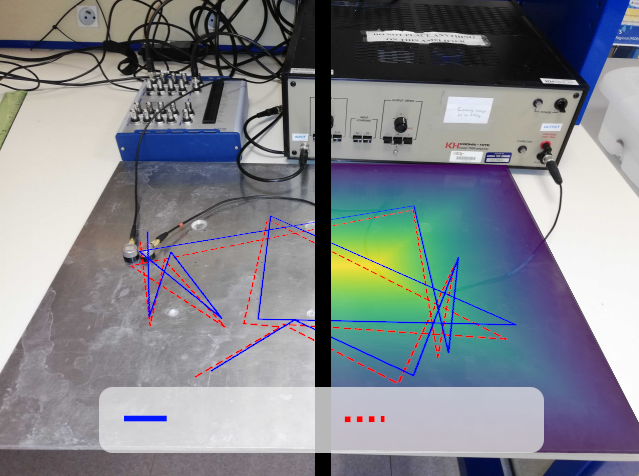}};
        \node[anchor=west] at (3.3,-0.92){\tiny Ground truth};
        \node[anchor=west] at (3.3,-1.08){\tiny trajectory};
        \node[anchor=west] at (4.8,-0.92){\tiny Estimated};
        \node[anchor=west] at (4.8,-1.08){\tiny trajectory};
        \node[below=0.1cm of bat,inner sep=0,outer sep=0] {\scriptsize (a) Proposed distance field};
        \node[below=0.1cm of real,inner sep=0,outer sep=0] {\scriptsize (b) Real-world echolocation};
    \end{tikzpicture}
    \caption{Illustration of the proposed continuous GP-based distance field compared with the naive smooth minimum approach solely using noisy observations of the surface (a). The high level of accuracy allows for applications such as echolocation shown in (b): real-world experimental setup with an ultrasonic emitter-receiver, and estimated trajectory.}
    \label{fig:teaser}
\end{figure}

A distance field (or distance function) is a function over a given space $\reals^N$ that maps a query point $\abscissavec$ with the distance $\dist$ to the nearest object/surface.
For some regular-shaped objects, the surface can be represented with a \emph{parametric} or \emph{implicit} function that equals zero on the surface.
The knowledge of such a function (eg., $ax + by + cz + w = 0$ for an infinite plane) may lead to a closed-form expression of the distance field.
In the general case, the surface function and the distance field are not explicitly known, and in the context of robotics, surfaces are often represented with discrete samples in the form of noisy point clouds (Fig.~\ref{fig:teaser}~(a)).
Distance-field-based frameworks have increased in popularity over the past decade.
Using a truncated distance function, KinectFusion \cite{kinectfusion} performs dense 3D reconstruction using an RGB-D camera. \acp{esdf} \cite{Helen2016signed} allowed for novel mapping and planning methods \cite{Voxblox,han2019fiesta,ortiz2022isdf}.
More recently, leveraging the computational power of GPUs, neural-network-based representations also exploit distance fields.
Several works build upon such representations for the purpose of \ac{slam} \cite{sucar2021imap,zhu2022nice_slam} and path planning \cite{driess2022learningSDF_planning,pantic2022NeRF_planning}.

\ac{gp} regression is a non-parametric, probabilistic interpolation method widely used for representing the environment.
In \cite{Microsoft}, surface reconstruction is performed with \acp{gpis} modelling the world as a continuous scalar field representing the signed distance to the surface close to the object.
Unfortunately, this representation does not allow for path planning/obstacle avoidance as the field converges to zero away from the surface.
In \cite{Wu2021}, the LogGPIS is introduced to remedy this issue by defining a distance field over the full space.
It has been shown that LogGPIS is suitable for applications like odometry, mapping, path planning, and surface reconstruction \cite{Wu2022log}.
These applications rely on the fact that the distance field is minimal on the surface (e.g., distance minimisation for scan registration), and maximal the furthest away from the surface (e.g., distance maximisation for obstacle avoidance).
However, LogGPIS does not provide a truly Euclidean distance field (c.f. Section~\ref{sec:background_dist}) and suffers from a trade-off between accuracy and interpolation abilities \cite{ivan2022online}.
These are limiting factors for specific applications that require accurate values of the Euclidean distance field such as echolocation.
Additionally, due to the non-linear transformation of the \ac{gp}-field, the variance prediction of LogGPIS fails to reliably model the field's uncertainty.
The proposed method addresses these issues with a more accurate representation of the distance field and by providing a proxy for its uncertainty.

Localisation and mapping based on range measurements is an active field of study.
It finds a wide range of applications such as acoustic room geometry reconstruction~\cite{Dokmanic2016, Krekovic2016}, underwater localisation and mapping~\cite{Ribas2007,cheung2019non}, and robotic inspection~\cite{Worley2020, Ouabi2021}.
Yet, recent works often rely on strong assumptions on the map.
For example, in~\cite{Dokmanic2016, Krekovic2016, Ouabi2021} the map is modelled with a fixed set of lines, preventing the possibility to capture richer information on the environment.
\ac{gp}-based distance fields are appealing in these applications as they can represent complex environments with non-parametric models.
The accuracy of the proposed distance field enables us to tackle the problems of echolocation and mapping based on \acp{ugw}.

To summarise, the contributions outlined in this work further the abilities of \ac{gp} regression through the derivation of a novel \ac{gp}-based distance field and a proxy for its uncertainty using solely point clouds as input, the thorough evaluation of the proposed distance field, and its integration into echolocation and mapping frameworks using \acp{ugw}.

\section{Background}

The design of the proposed distance field relies on \ac{gp} regression.
Accordingly, this section provides the required preliminary knowledge and explores existing distance field methods and their limits.

\subsection{Gaussian Process regression}

\ac{gp} regression is a probabilistic non-parametric method for interpolation.
Let us consider an unknown signal $h(\abscissavec) \in \reals$ with $\abscissavec \in \reals^{m}$, and $Q$ noisy observations $\ordinate_i$ defined as
\begin{align}
    \ordinate_i = h\left( \abscissavec_i \right) + \noise_i,\quad \text{where}\ \noise_i \sim \gaussian\left( 0, \sigma_y^2 \right),
\end{align}
where $i = (1,\cdots, Q)$.
The goal is to infer the distribution (mean and variance) of $h$ for any given input $\abscissavec$.

By modelling the signal $h$ as a \ac{gp} $h \sim \gp\left( 0, \kernel{}{\abscissavec}{\abscissavec'}\right)$, with $\kernelraw{}$ the covariance kernel function $\kernel{}{\abscissavec}{\abscissavec'} = \text{cov}\left(h(\abscissavec), h(\abscissavec')\right)$, one can express occurrences of $h$ as a multivariate Gaussian distribution
\begin{align}
    \begin{bmatrix}
        \ordinatevec \\ h(\abscissavec)
    \end{bmatrix}
    = 
    \gaussian\left(0,
    \begin{bmatrix}
        \kernelmatraw{\abscissamat\abscissamat} + \sigma_y^2\identitymat & \kernelvecraw{\abscissavec\abscissamat}^\top 
        \\
        \kernelvecraw{\abscissavec\abscissamat} & \kernel{}{\abscissavec}{\abscissavec}
    \end{bmatrix}\right)\,,
    \label{eq:multivariate_gaussian}
\end{align}
where $\ordinatevec = [\ordinate_1, \cdots, \ordinate_Q]^\top$, $\abscissavec$ is a query point, $\kernelvecraw{\abscissavec^*\abscissamat}=[ \kernel{}{\abscissavec}{\abscissavec_1},\cdots,\kernel{}{\abscissavec}{\abscissavec_Q}]$, and $\kernelmatraw{\abscissamat\abscissamat} = [\kernelvecraw{\abscissavec_1\abscissamat}^\top, \cdots,\kernelvecraw{\abscissavec_Q\abscissamat}^\top]$.
By conditioning \eqref{eq:multivariate_gaussian} with respect to the noisy observations, the mean and variance of $h(\abscissavec^*)$ are respectively computed as
\begin{align}
    \begin{aligned}
    \hat{h}(\abscissavec) = &\kernelvecraw{\abscissavec\abscissamat} \left(\kernelmatraw{\abscissamat\abscissamat} + \sigma_y^2\identitymat \right)^{-1} \ordinatevec,
    \label{eq:gp_inference}
    \\
    \sigma^2_{(\hat{h}(\abscissavec))} = &\kernel{}{\abscissavec}{\abscissavec}
    -\kernelvecraw{\abscissavec\abscissamat} \left(\kernelmatraw{\abscissamat\abscissamat} + \sigma_y^2\identitymat \right)^{-1} \kernelvecraw{\abscissavec\abscissamat}^\top.
    \end{aligned}
\end{align}

\subsection{Distance fields}
\label{sec:background_dist}
Considering a surface $\surface$ in Euclidean space $\reals^m$, let us define the distance field $\dist(\abscissavec)$ with $\abscissavec \in \reals^m$ as a scalar-valued continuous function that represents the shortest distance between the input $\abscissavec$ and the surface $\surface$.
Such a function is a solution to the Eikonal equation
\begin{align}
    \lvert \gradient \dist(\abscissavec) \lvert = 1\quad\text{with}\quad \dist(\abscissavec) = 0\ \iff \abscissavec \in \surface.
    \label{eq:eikonal}
\end{align}
Unfortunately, as per its non-linear nature, \eqref{eq:eikonal} does not possess a known general closed-form solution.
The aim of this work is to estimate the Euclidean distance $\distfield$ given a finite set of points on the surface $\abscissavec_i \in \surface$ with $i=(1,\cdots,Q)$.
It is a challenging task in the sense that only samples of $\dist(\abscissavec) = 0$ are provided.
Implicitly, this task requires representing the surface in a continuous manner to enable the interpolation between noisy measurements and provide an accurate estimate of the distance to the surface.
In this section, we explore a few existing continuous and differentiable distance field approximations.

\subsubsection{Smooth-minimum}
\label{sec:smooth_min}
A naive way to address the distance function estimation problem is to compute the distance between the query point $\abscissavec$ and each of the surface observations $\abscissavec_i$ and take the minimum of these distances.
Unfortunately, the derivative of such an approach is not continuous.
However, one can leverage the \emph{smooth minimum} function defined as $\frac{\left( \sum_{i=1}^{Q}\Vert \abscissavec - \abscissavec_i \Vert\exp(\lambda\Vert \abscissavec - \abscissavec_i \Vert)\right)}{\left( \sum_{i=1}^{Q}\exp(\lambda\Vert \abscissavec - \abscissavec_i \Vert)\right)}$
.
While it is suitable for use in optimisation cost functions as it is differentiable, it does not account for any observation noise.
Furthermore, as illustrated in Fig.~\ref{fig:teaser}, the smooth-minimum function does not interpolate the surface between the discrete observations.

\subsubsection{GPIS}
Originally the \ac{gpis} \cite{Microsoft} represents the surface with the zero crossing of a \ac{gp}-modelled scalar field.
It is achieved by arbitrarily fixing the value of the surface observation to zero and adding positive and negative virtual observations inside and outside the closed surface.
The interpolation abilities of \ac{gp} regression allow \ac{gpis} to interpolate the surface between discrete measurements.
In \cite{martens2017geometric}, with the use of linear operators \cite{Sarkka2011} and observations of the normal vectors, the scalar field corresponds to the distance field close to the surface but converges to zero further away.
It results in a scalar field that is not monotonic with respect to the true Euclidean distance to the surface.
This is a significant drawback that prevents the use of GPIS for applications like scan registration or path planning.

\subsubsection{LogGPIS}
\label{sec:background_loggpis}
Motivated by heat-based methods \cite{Crane2013heat}, LogGPIS \cite{Wu2021} models the propagation of heat in space at time $t$ with a zero-mean \ac{gp} scalar field $v_t(\abscissavec) \sim \mathcal{GP}(0,\kernel{v}{\abscissavec}{\abscissavec'})$.
It corresponds to the assumption that the surface's temperature is maintained to one and that the rest of the space has an initial temperature of zero.
The time $t$ is directly deduced from the kernel's lenghtscale $l$.
According to Varadhan's results \cite{Varadhan1967}, the limit $\lim_{t\to 0} \{-\sqrt{t}\ln\left(v(\mathbf{x}, t)\right)\}$ is equal to the distance from the surface.
Using a fixed lengthscale and a specific kernel, the resulting distance field is the solution of a regularised version of the Eikonal equation.
Unfortunately, this leads to a trade-off between accuracy and interpolation abilities: the limit in Varadhan's equation requires a small lengthscale to approximate the true distance to the surface, but a small lengthscale removes the correlation between the \ac{gp} observations, reducing the ability of the method to interpolate non-observed parts of the surface.
Additionally, regardless of the kernel's lengthscale, LogGPIS does not represent a truly Euclidean distance field of space.
Considering the simple case of using only one observation $\abscissavec_0$ of the surface and the \matern $3/2$ kernel, the heat is analytically expressed as $v_t(\abscissavec) = (1+\frac{\sqrt{3}d}{l})\exp(-\frac{\sqrt{3}d}{l})$.
Therefore, one can compute the error between the estimated distance and the ground truth as $-\frac{l}{\sqrt{3}}\log\left(v_t\left(\abscissavec\right)\right) - d = -\frac{l}{\sqrt{3}} \log(1+\frac{\sqrt{3}d}{l})$.
The growth of the absolute error with respect to the distance expends to multi-observation scenarios as empirically shown later in our experiments.

\section{Method}
\label{sec:method}
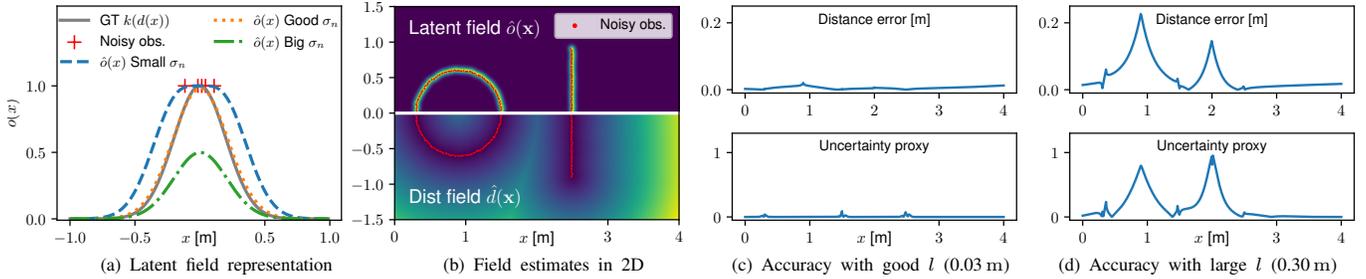
\begin{figure*}
   \centering
    \def\legendsize{\scriptsize}
    \def\horispacing{0.1cm}
    \def\vertspacing{-0.15cm}
   \begin{tikzpicture}
        \node (img)        {\includegraphics[clip, width=0.99\textwidth, trim=0.3cm 0.5cm 0.0cm 0.3cm]{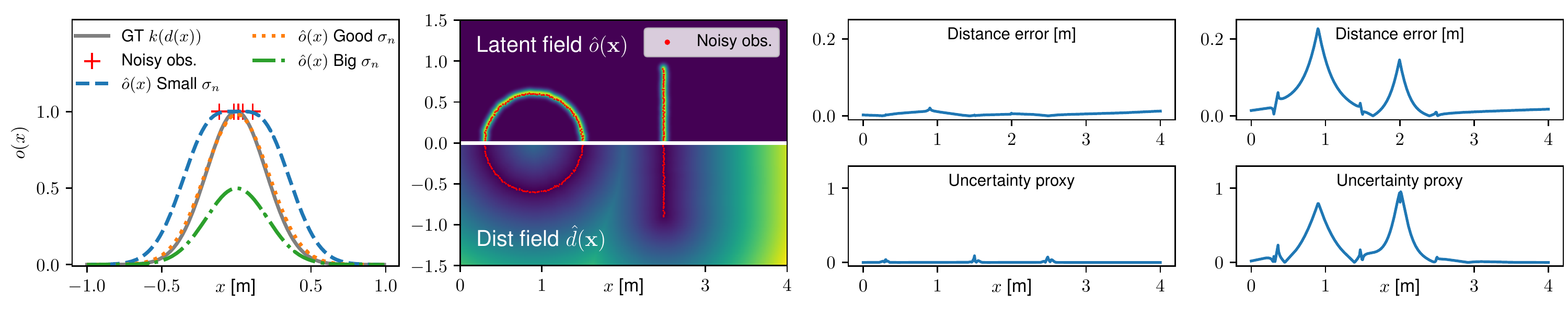}};
        \node [text width=4cm, minimum width = 4cm, align=center, below=\vertspacing of img, xshift = -6.2cm] (legenda) {\legendsize(a) Latent field representation};
        \node [text width=4cm, minimum width = 4cm, align=center, right=\horispacing of legenda] (legendb) {\legendsize(b) Field estimates in 2D};
        \node [text width=4cm, minimum width = 4cm, align=center, right=\horispacing of legendb] (legendc) {\legendsize(c) Accuracy with good $\lengthscale$ (0.03$\units{m}$)};
        \node [text width=4cm, minimum width = 4cm, align=center, right=\horispacing of legendc] (legendd) {\legendsize(d) Accuracy with large $\lengthscale$ (0.30$\units{m}$)};
   \end{tikzpicture}
   \vspace{-0.7cm}
   \caption{Illustration of the proposed GP-based distance field and the effect of hyperparameter values. (a) shows in 1D the GP-modelled latent field $\occ(\abscissavec) = \kernelraw{}(\dist(\abscissavec))$. Provided appropriate hyperparameters, the inference matches the ground truth model. (b) is an example of field inference in 2D. (c) and (d) represent the absolute error and uncertainty proxy along the white line in (b) with two different kernel lengthscales.}
   \label{fig:model_and_uncertainty}
\end{figure*}

The proposed method comes from the observation that applying an unscaled, monotonic, stationary, isotropic kernel function\footnote{Such kernel function can be expressed as a function of the distance between the two input arguments: $\kernel{}{\abscissavec}{\abscissavec'}\to\kernelraw{}(\Vert \abscissavec - \abscissavec' \Vert)$.} $\kernelraw{}(\dist)$ to the distance field generated by a single point in $\reals^N$, results in exactly the same scalar field as performing \ac{gp} regression using a zero-mean model and a single noiseless observation equal to $\kernelraw{}(0) = 1$.
Further, leveraging the probabilistic nature of \ac{gp} regression and given the appropriate hyperparameters, we can obtain a very similar scalar field using noisy measurements (c.f., Fig.~\ref{fig:model_and_uncertainty}).
Accordingly, we define $\occ(\abscissavec) = \kernelraw{o}(\dist(\abscissavec))$ a latent scalar field modelled with a zero-mean \ac{gp}: $\occ(\abscissavec)\sim\gp\left( 0, \kernel{\occ}{\abscissavec}{\abscissavec'}\right)$.
Note that provided a set of surface observations $\abscissavec_i$ (with $\kernelraw{o}(\dist(\abscissavec_i)) = 1$), the field $\occ(\abscissavec)$ can be seen as a smooth representation of the occupancy space, which is similar to the heat field of LogGPIS.
By using \eqref{eq:gp_inference}, we can infer the latent field at any point $\abscissavec \in \reals^N$:
\begin{align}
    \hat{\occ}(\abscissavec)& = \kernelvecraw{\abscissavec\abscissamat} \left(\kernelmatraw{\abscissamat\abscissamat} + \sigma_{\occ}^2\identitymat \right)^{-1} \mathbf{1},
    \label{eq:occ_inference}
\end{align}
with $\sigma_o$ a hyperparameter discussed in Section~\ref{sec:uncertainty}.
The proposed distance field is obtained by applying the inverse of the kernel function to the latent field:
\begin{align}
    \distfield(\abscissavec) = \revert\left(\hat{\occ}\left(\abscissavec\right)\right)\ \ \text{with}\ \
    \revert\left( \kernelraw{}(\Vert \abscissavec - \abscissavec' \Vert) \right) \triangleq \Vert \abscissavec - \abscissavec' \Vert.
    \label{eq:revert_def}
\end{align}
We denote $\revert$ as the \emph{reverting} function of our latent \ac{gp} field.

In order to use the distance field in any optimisation-based framework (odometry, planning, etc) or for shape reconstruction, we need to infer the gradient/normal of the field.
With the use of linear operators \cite{Sarkka2011}, it is also possible to recover the gradient $\nabla\occ$ of the latent field with its associated covariance $\Sigma_{\nabla\occ}$ as:
\begin{align}
    \nabla\hat{\occ}(\abscissavec)& = \nabla\kernelvecraw{\abscissavec\abscissamat}\left(\kernelmatraw{\abscissamat\abscissamat}+\sigma_n\identitymat\right)^{-1}\mathbf{1}
    \\
    \Sigma_{\nabla\hat{\occ}}(\abscissavec)& = \nabla\kernelraw{\abscissavec\abscissavec}\nabla 
    -\nabla\kernelvecraw{\abscissavec\abscissamat} \left(\kernelmatraw{\abscissamat\abscissamat}+\sigma_n\identitymat\right)^{-1}\left(\nabla\kernelvecraw{\abscissavec\abscissamat}\right)^\top.
    \nonumber
\end{align}
The value of the gradient and its uncertainty are also used as part of the uncertainty estimation detailed in Section~\ref{sec:uncertainty}.

\subsection{Kernels, reverting functions}

\label{sec:kernels}
This subsection presents the reverting functions associated with a few commonly-used kernels that are at least once-differentiable the \ac{rq} kernel, the \ac{se} kernel, and the $\nu=3/2$ \matern kernel.
In their unscaled isometric form, these kernels depend on the Euclidean distance between the two input vectors.
Using $\dist = \Vert \abscissavec - \abscissavec' \Vert$, Table~\ref{tab:kernels} shows the different kernels as well as their reverting function with $\lengthscale$ the kernel's lengthscale.
In the case of the \ac{rq} and \ac{se} it is straightforward to find the reverting function with simple algebraic manipulations.
Unfortunately, the $\nu=3/2$ \matern kernel does not possess a known, closed-form reverting function.
In this scenario, the reverting ``function" can be formulated as a single-value non-linear optimisation problem.

\begin{table}[]
    \centering
    \begin{scriptsize}
    \newcolumntype{Y}{>{\centering\arraybackslash}X}
    \newcolumntype{C}[1]{>{\centering\let\newline\\\arraybackslash\hspace{0pt}}m{#1}}
    \begin{tabularx}{1\linewidth}{C{1cm} || Y | Y}
        & \textbf{Covariance kernel} ($\kernelraw{\occ}(\dist)$) & \textbf{Reverting function} ($\revert(\occ)$)
        \\ \hline \hline
        Rational quadratic & $\left(1+\frac{\dist^2}{2\alpha\lengthscale^2}\right)^{-\alpha}$ & $\sqrt{2\alpha\lengthscale^2\left(\occ^{-\frac{1}{\alpha}}-1\right)}$
        \\
        \hline
        Square exp. & $\exp\left(-\frac{\dist^2}{2\lengthscale^2}\right)$ & $\sqrt{-2\lengthscale^2\log\left(\occ\right)}$
        \\
        \hline
        \matern $\nu=3/2$ &$\left(1+\frac{\sqrt{3}\dist}{\lengthscale}\right)\exp\left(-\frac{\sqrt{3}\dist}{\lengthscale}\right)$ & $\underset{\dist}{\text{argmin}}\ \lVert \occ - \kernelraw{\occ}\left(\dist\right)\rVert^2$
    \end{tabularx}
    \end{scriptsize}
    \caption{Covariance kernels and associated reverting functions}
    \label{tab:kernels}
\end{table}

The value of the lengthscale depends on the data at hand as it corresponds to the typical distance at which the data points are correlated.
Intuitively, $\lengthscale$ represents the ``level of interpolation" as it characterises the ``area of influence" of each of the observed points on the surface.
We show in Fig.~\ref{fig:model_and_uncertainty} the impact of the lengthscale on the distance field estimate.
While it is safe to define the lengthscale $\lengthscale$ empirically equal to one and a half times the typical distance between neighbour data points of the same surface, future work will explore the systematic learning of this hyperparameter from the data.

\subsection{Uncertainty modelling}

\label{sec:uncertainty}
Let us first discuss the uncertainty of the noisy observations.
Unlike standard \ac{gp} regression where the noise is modelled on the value of the observations, here the uncertainty is on the location of the observation.
This is addressed in \cite{mchutchon2011gaussian} by adding a corrective term to the observations' covariance based on the derivative of the \ac{gp} mean.
Unfortunately, the derivative of our latent field's mean is null.
We then propose to learn the corrective term $\sigma_n$ (corresponding to the measurement noise in standard \ac{gp} regression) by minimising the Mahalanobis distance between the \ac{gp} posterior given a typical set of noisy observations (real or simulated from expected noise specification and typical point density) and the desired latent field model.
Formally, considering a set of noisy observations $\abscissamat$ of a single surface and a set of query points $\abscissamat_q$ (from a regular grid around the surface) with their associated ideal latent space values $\occ_q(\abscissamat_q) = \kernelraw{}(\dist(\abscissamat_q))$, we can define the optimisation problem as 
\begin{equation}
    \sigma_n^* = \underset{\sigma_n}{\mathrm{argmin}} (\mathbf{\occ}_q - \hat{\mathbf{\occ}})^\top\Sigma_{\hat{\occ}}^{-1}(\mathbf{\occ}_q - \hat{\mathbf{\occ}}),
    \nonumber
\end{equation}
with $\hat{\mathbf{\occ}}$ and $\Sigma_{\hat{\occ}}$ obtained with \eqref{eq:occ_inference} and \eqref{eq:gp_inference}.
Fig.~\ref{fig:model_and_uncertainty} illustrates the results of such an optimisation approach (the grey line ``GT" is the desired latent field).

Regarding the output of our method $\distfield$, it is important to note that due to the non-linear nature of the reverting function, the proposed distance field is not a \ac{gp}.
While one can propagate the variance of the latent field through the reverting function using its Jacobian as done in LogGPIS, this does not accurately represent the uncertainty of the estimated distance field as it rapidly diverges towards infinity.
We propose to use the discrepancy between the derivative of the proposed latent field model ($\occ(\abscissavec) = \kernelraw{}(\dist(\abscissavec))$) and the gradient of the inferred field ($\hat{\occ}(\abscissavec)$) as a proxy for the distance uncertainty.
This uncertainty proxy is defined as the Mahalanobis distance between $\frac{\partial \vert \kernelraw{}(\dist)\vert}{\partial \dist}\mid_{d = \hat{d}}$ and $\Vert\nabla \hat{\occ}(\abscissavec) \Vert$ leveraging the covariance $\Sigma_{\hat{\occ}}(\abscissavec)$ propagated through the norm.
The value of the proposed uncertainty proxy is shown in Fig.~\ref{fig:model_and_uncertainty}~(b-d).

\subsection{Discussion and further developments}
In the general scenario with noisy measurements, despite efforts to provide good values for the hyperparameters as illustrated in Fig.~\ref{fig:model_and_uncertainty}~(a), there is no guarantee that the field $\hat{\occ}$ is exactly one ($\implies \distfield = 0$) on the surface.
Consequently, for the sake of shape reconstruction (the original motivation of \ac{gpis}), leveraging only the latent or distance field is not sufficient.
To determine exactly where the surface is, one needs to analyse the derivatives of the latent field along the normal.
By considering the probability $p(\hat{\occ}>thr, \Vert \nabla\hat{\occ}\Vert <\epsilon \mid \abscissamat)$, a probabilistic and continuous version of the marching cubes algorithm \cite{lorensen1987marching} could be developed.
In our distance field implementation, we cap the latent field to one before applying the reverting function\footnote{Note that our uncertainty proxy still provides information about these imperfect estimates.}.
On the other extreme, when querying far from the surface (relative to the kernel's lenghtscale), the success of the proposed field can be hindered due to limited machine precision.
Accordingly, inferences of $\hat{\occ}$ that are not strictly positive are regarded as failure cases.

Interestingly, LogGPIS turns out to be a close-range approximation of the reverting function concept\footnote{Approximating the \matern kernel used in LogGPIS ($\nu = 3/2$) for small distances $\lim_{\dist \to 0} \left(1+\frac{\sqrt{3}\dist}{\lengthscale}\right)\exp\left(-\frac{\sqrt{3}\dist}{\lengthscale}\right) \approx \exp\left(-\frac{\sqrt{3}\dist}{\lengthscale}\right)$, the associated reverting function would be $\tilde{\revert}(\occ) = - \frac{\lengthscale}{\sqrt{3}}\log(\occ)$. This corresponds to the logarithm operation introduced in LogGPIS.}.
Unlike LogGPIS, the proposed formulation does not rely on a specific kernel.
However, not all kernels can be used due to the need for the derivatives in the uncertainty proxy and the downstream applications.
Thus, the kernel needs to be at least once-differentiable (unlike \matern 1/2 or Whittle kernels).
Section~\ref{sec:experiments} provides empirical-based guidelines about the choice of the kernel.
Nonetheless, additional work is needed to understand better the impact of the kernel choice on the physical meaning of the latent field and the properties of the resulting distance estimates.

\section{Applications}

\begin{figure}
    \centering
    \begin{tikzpicture}
        \node {\includegraphics[clip, width=\columnwidth, trim= 0cm 0cm 0cm 0cm]{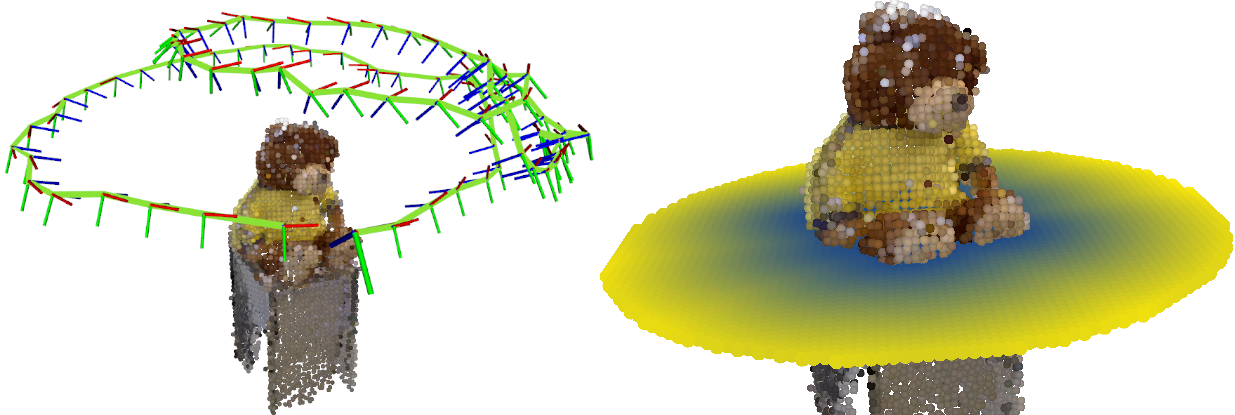}};
        \node[align = left, text width = 2.8cm, minimum width=3cm, execute at begin node=\setlength{\baselineskip}{1.5ex}] at (-2.9,-0.8) {\scriptsize \textbf{Traj. RMSE}:\\$0.021\units{m}$\\$2.09^\circ$};
    \end{tikzpicture}
    \vspace{-0.65cm}
    \caption{Results of RGB-D odometry and mapping with the proposed distance field (slice visible on the right). Dataset: Freiburg3/teddy \cite{Sturm2012}.}
    \label{fig:odometry}
\end{figure}

As both LogGPIS \cite{Wu2022log} and the proposed method use monotonic functions over a similar \ac{gp}-modelled latent field, one can expect both fields to have colocated local minima and maxima.
Thus, all the applications demonstrated in \cite{Wu2022log} (depth-camera odometry and mapping, path planning, surface reconstruction) can be performed with the proposed method.
As an example, we released a real-time open-source implementation of a depth-camera odometry and mapping pipeline based on the proposed distance field\footnote{https://github.com/UTS-CAS/gp\_odometry} (c.f., Fig.~\ref{fig:odometry}).
Due to the space limitation, we only present novel uses of \ac{gp}-based distance fields such as echolocation and mapping with \ac{ugw} sensing systems.
Let us first introduce the \ac{ugw} measurement model.

\subsection{Ultrasonic Guided Waves measurement model}
\label{subsec:ugw_model}

\begin{figure}
   \centering
    \def\legendsize{\scriptsize}
   \begin{tikzpicture}
        \node (imgsrc)        {\includegraphics[width=0.59\columnwidth]{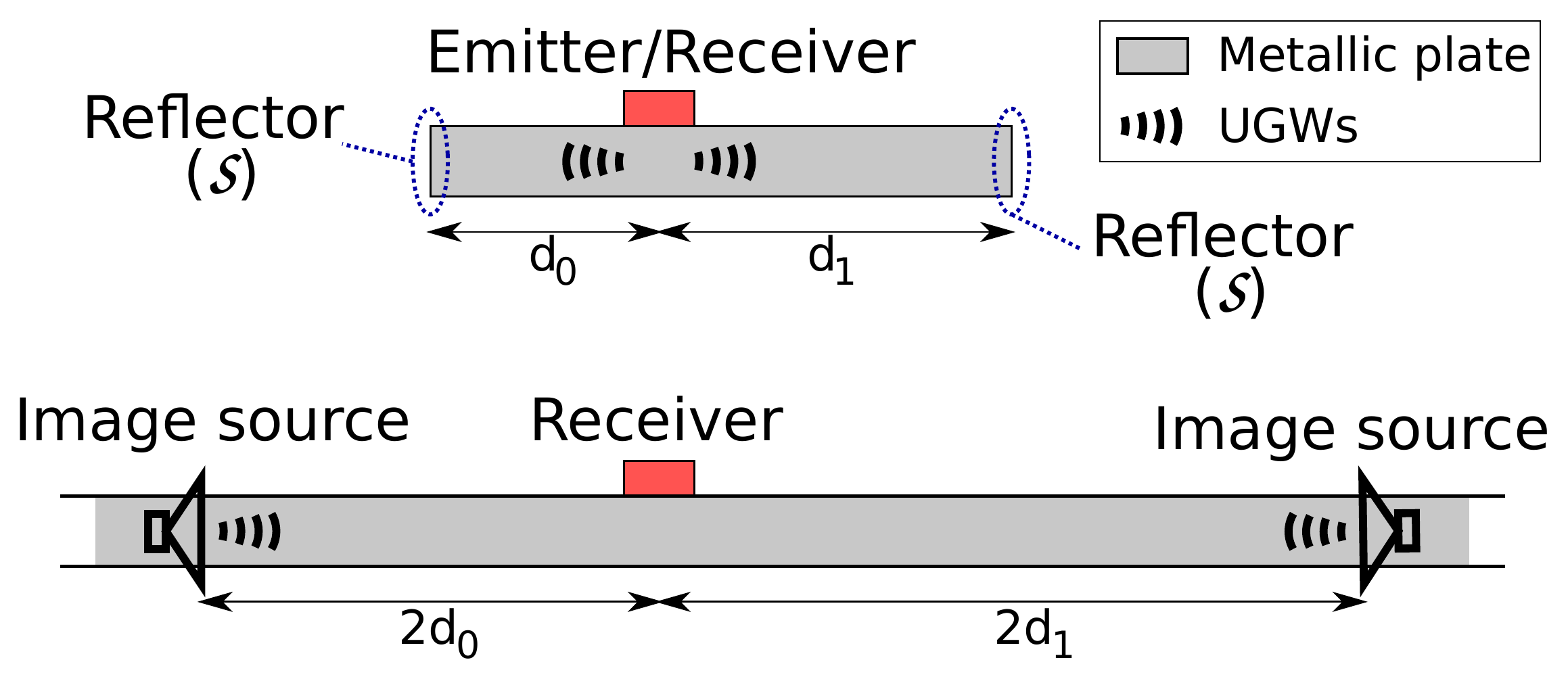}};
        \node [right=0cm of imgsrc] (envelop) {\includegraphics[width=0.39\columnwidth]
        {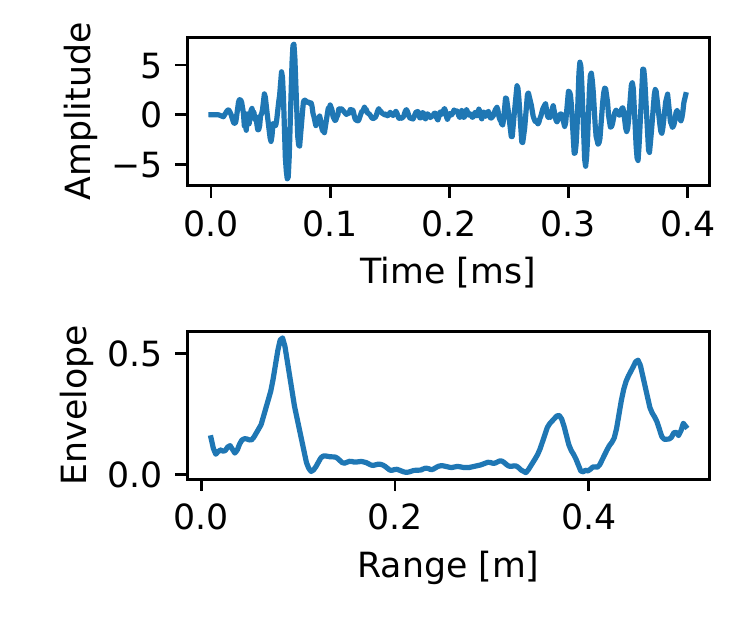}};
        \node[align=center, below=0.05cm of imgsrc]{\legendsize (a) Emitter/receiver and \emph{image source} representation};
        \node[align=center, below=-0.27cm of envelop]{\legendsize (b) Real signal and envelope};
   \end{tikzpicture}
   \vspace{-0.7cm}
   \caption{Illustration of the emitter/receiver setup and the equivalent \emph{image source} representation when considering only first-order echoes (a), and example of an acoustic measurement acquired on an aluminium plate with the corresponding envelope signal (b).}
   \label{fig:example_signal}
\end{figure}

Given a co-located emitter-receiver, at every measurement step $i$, a time signal $s(t)$ is pulsed by the emitter to create an \ac{ugw} that propagates radially around the emitter, \textit{inside} the structure's material.
Simultaneously, the receiver collects the measurement $z_i(t)$ that contains the ultrasonic echoes due to reflections of the excited wave on structural features (the boundaries $\surface$ of a metal panel in our case study).
These measurements can be modelled by relying on the \emph{image source} model, which states that the reception of any echo can be interpreted as a signal originating directly from a fictive image source (cf. Fig.~\ref{fig:example_signal}(a)).
The image sources' positions depend on the position of the actual emitter and the geometry of the surface $\surface$.
Considering isotropic material, the propagation is only a function of time and of the distance between the receiver and the image source resulting in the following measurement model: $\tilde{z}_i(t) = \sum_{\boldsymbol{\rho} \in \mathcal{I}(\pos{i}, \surface)} g(||\boldsymbol{\rho} - \pos{i}{}||, t) \ast s(t) + n_i(t)$, where $\pos{i}{}$ is the emitter position, $\mathcal{I}(\pos{i}, \surface)$ is the set of the image source positions for a surface $\surface$ and a real source position $\pos{}{}$, $g(||\boldsymbol{\rho}- \pos{i}||, t)$ is the acoustic transfer function of the propagation medium\footnote{A standard propagation model for an \ac{ugw} propagating in a metal panel is $\hat{g}(r, \omega) \approx e^{- j k(\omega)r} / \sqrt{k(\omega)r}$, where $k(\omega)$ is the wavenumber of the major acoustic mode. More details can be found in~\cite{Su2009}.}, $n(t)$ is an additive Gaussian noise term that we assume temporally and spatially white, and the symbol~$\ast$ denotes the convolution operation.

Let us generate a correlation signal between the measurement and the model to assess the likelihood that a \textit{single} acoustic reflection occurred at any distance $d$ from the emitter-receiver with:
\begin{equation}
    z'_i(d) =  \frac{\langle z_i(t), \hat{z}(d, t)\rangle}{ \sqrt{\langle z_i(t), z_i(t) \rangle \langle \hat{z}(d, t), \hat{z}(d, t) \rangle} },
    \label{correlation_signal}
\end{equation}
where $\hat{z}(d, t) = \hat{g}(2 d, t) \ast s(t)$ is the expected signal containing an echo due to a reflection at a distance $d$, and $\langle., .\rangle$ denotes the usual scalar product for time-continuous signals.
We subsequently retrieve the envelope of the correlation signals with $e_i(d) = \left| z'_i(d) + j \mathcal{H}(z'_i)(d) \right|$, with $\mathcal{H}$ denoting the Hilbert transform operator.
The envelope signal ranges between 0 and 1, and it presents a local maximum in $\dist$ if there is indeed a reflector at such distance, as illustrated in~Fig.~\ref{fig:example_signal}(b).
The figure shows a distinguishable first echo at $0.08\units{m}$ corresponding to the closest boundary and later echoes corresponding to the other edges of the plate.

To summarise, from a real measurement and the propagation model, the envelope signal allows us to assess the likelihood of a reflector being present at any distance $\dist$.
Accordingly, considering the map (known or estimated) of the reflective features in the environment in the form of a point cloud and using the proposed \ac{gp}-based distance field, it is possible to evaluate the envelope signal at any position in space to quantify the likelihood of a measurement being captured at such location.

\subsection{Echolocation}
We aim at estimating the system's position $\pos{i}{}$ by relying on ultrasonic measurements, noisy odometry information, and given the map (the plate boundaries $\surface$) in the form of a point cloud $\abscissamat$ using a particle filter as in~\cite{ouabi2021monte}.
Such filters usually provide satisfactory solutions to the localisation problem when the dynamic and observation models are non-linear, and the process and measurement noises are non-Gaussian~\cite{Thrun2005}.
Yet, the method described in~\cite{ouabi2021monte} can only be applied to metal panels with a rectangular shape.
Hence, we propose a modification to the calculation of the particles' weights to make the approach applicable to arbitrary surface geometries, using the proposed \ac{gp}-based distance field. 

The localisation approach relies on the aforementioned envelope signals $e_i(\dist)$ to determine the likelihood of each measurement and update the weights $w_i^{n}$ of the particles as follows: $w^{n}_i = \eta \exp \left( \beta e_i( \distfield(\pos{i}{p}) ) \right) w^{n}_{i-1}$,
where $\distfield(\pos{i}{n})$ is the GP-based distance of the $n$-th particle to the closest point on the surface $\surface$ computed with \eqref{eq:occ_inference} and the reverting function \eqref{eq:revert_def}, $\eta$ is the normalization factor, and $\beta$ is a positive parameter.
The filtering process consists of the succession of motion and measurement updates with regular resampling of the particles after a fixed number of steps (5 in our experiments).
The filter's output corresponds to the mean coordinates of the 25\% best particles (highest weights).

\subsection{Mapping}

For the purpose of mapping, the goal is to estimate the continuous distance field given the system's localisation and the \ac{ugw} measurements.
While the \ac{ugw} measurements contain information about the distance to the closest surface (peaks in the envelope $e_i(\dist)$), this scenario is different from the depth-camera/lidar mapping of \cite{Wu2022log} as there is no bearing information available.

Similarly to the echolocation scenario, we use the envelope signals $e_i(\dist)$ as proxies to quantify the likelihood of the measurements given the state variables.
In the context of echolocation, the map points $\abscissamat$ are known and the measurement positions $\pos{i}{}$ are estimated whereas for mapping it is the opposite: the sensing positions $\pos{i}{}$ are known and the state variables consist of a set of virtual observations $\abscissamat$ of the map (same $\abscissamat$ as in the \ac{gp} formulation in \eqref{eq:occ_inference}).
The maximisation of the envelope measurements is formulated as a non-linear least-square optimisation problem
\begin{align}
    \underset{\abscissamat}{\mathrm{argmin}} \left(\sum_{i=1}^{N} \Vert 1\text{-}e_i(\distfield( \pos{i}{}, \abscissamat) ) \Vert^2\! + \alpha\sum_{i=2}^{Q} \Vert \abscissavec_i\text{-}\abscissavec_{i\text{-}1} \Vert^2\right),
    \label{eq:envelop_mapping}
\end{align}
with $\alpha\sum_{i=2}^{Q} \Vert \abscissavec_i-\abscissavec_{i-1} \Vert^2$ a naive regularisation term to prevent under-constrained state variables in case virtual observation estimates are located in non-observed areas.
While being conceptually simple, solving \eqref{eq:envelop_mapping} is not an easy task due to the many extrema present in the envelope signals (c.f. Fig~\ref{fig:example_signal}(b)).
To provide a decent initial guess, we perform the mapping method considering only a single peak in each envelope signal by detecting the distances $\dist_i$ of the peaks corresponding to the first echoes:
\begin{align}
    \underset{\abscissamat}{\mathrm{argmin}} \left(\sum_{i=1}^{N} \Vert \dist_i \text{-} \distfield\left( \pos{i}{}, \abscissamat \right) \Vert^2\!+ \alpha\sum_{i=2}^{Q} \Vert \abscissavec_i\text{-}\abscissavec_{i\text{-}1} \Vert^2\right).
    \label{eq:dist_mapping}
\end{align}
Solving for \eqref{eq:dist_mapping} and then \eqref{eq:envelop_mapping} with the Trust Region Reflective algorithm allows for the estimation of the virtual observations therefore of the continuous occupancy field $\hat{\occ}$ and the associated distance field $\distfield$.
Note that the number of virtual observations should comply with the kernel's lengthscale and the length/area of the surface (i.e., enough points to cover~$\surface$).

\section{Experiments}

\label{sec:experiments}

\subsection{Distance field}
\label{sec:exp_dist_field}
\begin{table}[]
    \centering
    \begin{small}
    \newcolumntype{Y}{>{\centering\arraybackslash}X}
    \newcolumntype{C}[1]{>{\centering\arraybackslash}p{#1}}
    \begin{tabularx}{1\linewidth}{C{2.2cm} | Y  Y  Y}
$\quad\quad\quad\quad\quad$ Method & RMSE [m] close-range & RMSE [m] far-range & Coverage ratio\\
\hline
LogGPIS & 0.0132 & 0.0708 & 1.0 \\
Smooth min & 0.0190 & \underline{0.0104} & 1.0 \\
Ours SE & \textbf{0.0076} & \textbf{0.0018} & 0.3350 \\
Ours Matern 3/2 & \underline{0.0082} & 0.0260 & 1.0 \\
Ours RQ & \textbf{0.0076} & 0.0121 & 1.0 \\
    \end{tabularx}
    \end{small}
    \caption{Average RMSE (100 simulated environments $\times$ 40k samples)}
    \label{tab:rmse_comparison}
\end{table}

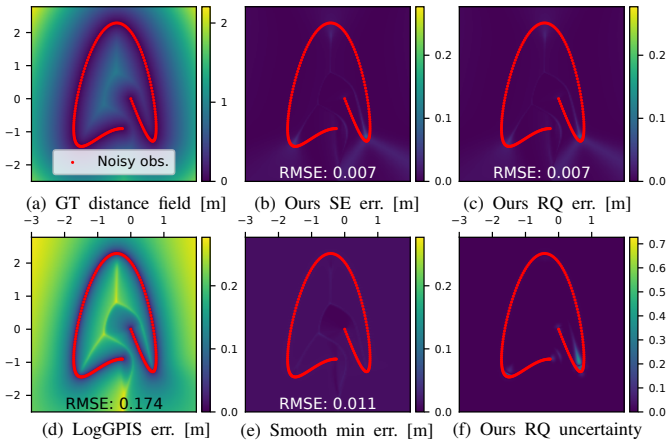
\begin{figure}
    \centering
    \def\vertspace{2.8cm}
    \def\horispace{-0.2cm}
    \def\legenddist{0.05cm}
    \def\legendsize{\scriptsize}
    \begin{tikzpicture}
     i     \node[inner sep=0,outer sep=0] (img) {\includegraphics[clip, trim=0cm 0.5cm 0cm 0.1cm, width=\columnwidth]{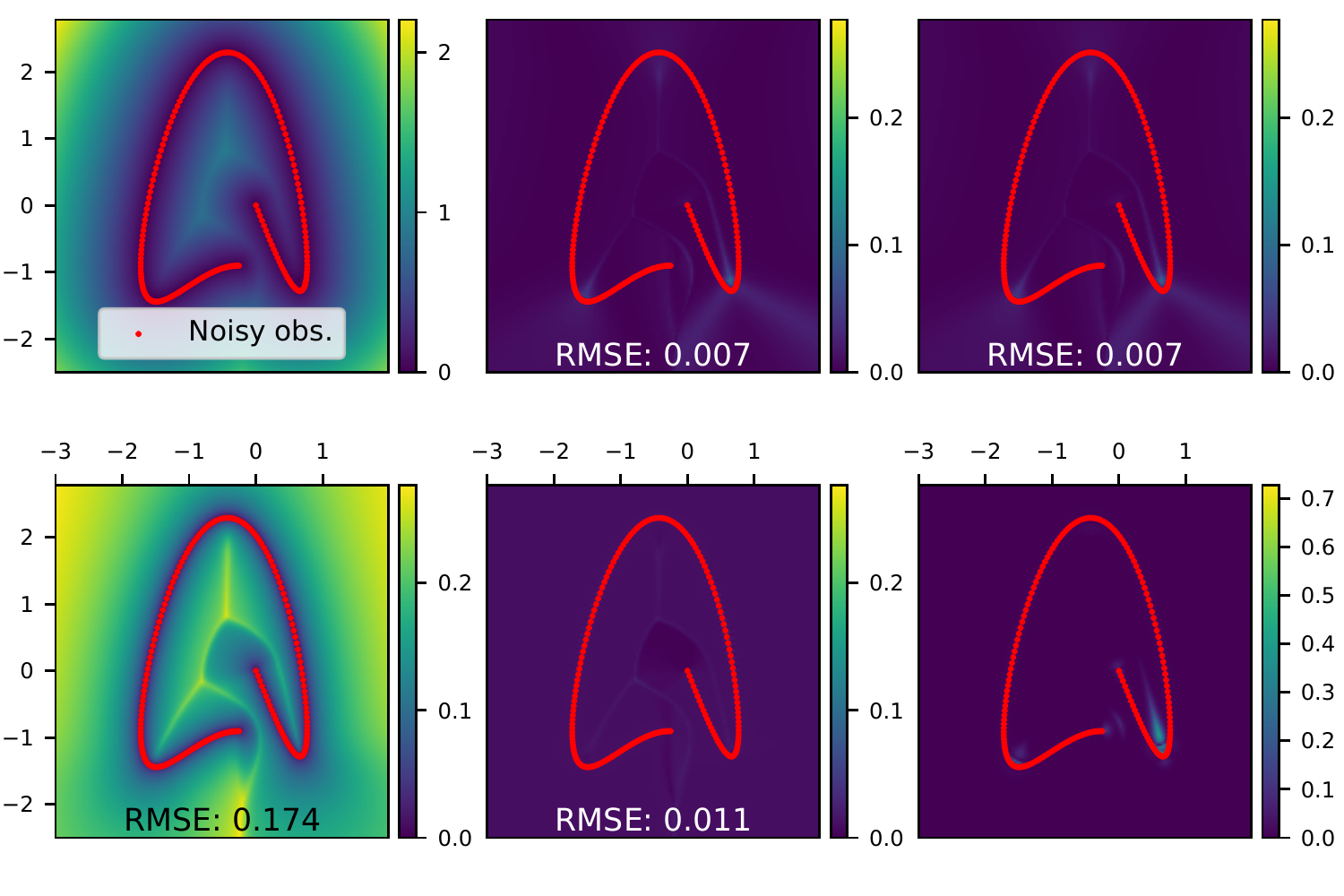}};

        \node[align = center, below=\legenddist of img,inner sep=0,outer sep=0, text width = 3cm, minimum width=3cm, xshift = -2.8cm] (d) {\legendsize (d) LogGPIS err. [m]};
        \node[align = center, above=\vertspace of d,inner sep=0,outer sep=0, text width = 3cm, minimum width=3cm] (a) {\legendsize (a) GT distance field [m]};
        \node[align = center, right=\horispace of a,inner sep=0,outer sep=0, text width = 3cm, minimum width=3cm] (b) {\legendsize (b) Ours SE err. [m]};
        \node[align = center, right=\horispace of b,inner sep=0,outer sep=0, text width = 3cm, minimum width=3cm] (c) {\legendsize (c) Ours RQ err. [m]};
        \node[align = center, right=\horispace of d,inner sep=0,outer sep=0, text width = 3cm, minimum width=3cm] (e) {\legendsize (e) Smooth min err. [m]};
        \node[align = center, right=\horispace of e,inner sep=0,outer sep=0, text width = 3cm, minimum width=3cm] (f) {\legendsize (f) Ours RQ uncertainty};
    \end{tikzpicture}
    \vspace{-0.6cm}
    \caption{Accuracy comparison of the proposed distance field, LogGPIS \cite{Wu2021}, and the smooth minimum function. (f) corresponds to the proposed uncertainty proxy.}
    \label{fig:kernel comparison}
\end{figure}

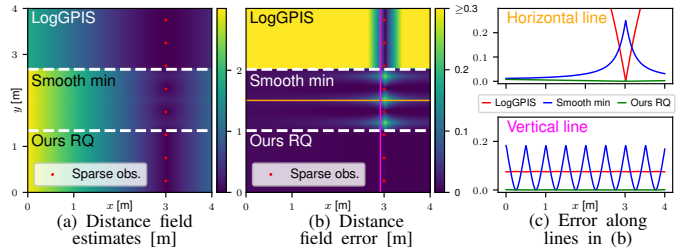
\begin{figure}
    \centering
    \def\horispace{0.08cm}
    \def\legenddist{-0.08cm}
    \def\legendsize{\scriptsize}
    \begin{tikzpicture}
     i     \node[inner sep=0,outer sep=0] (img) {\includegraphics[clip, trim=0.3cm 0.2cm 0cm 0.3cm, width=\columnwidth]{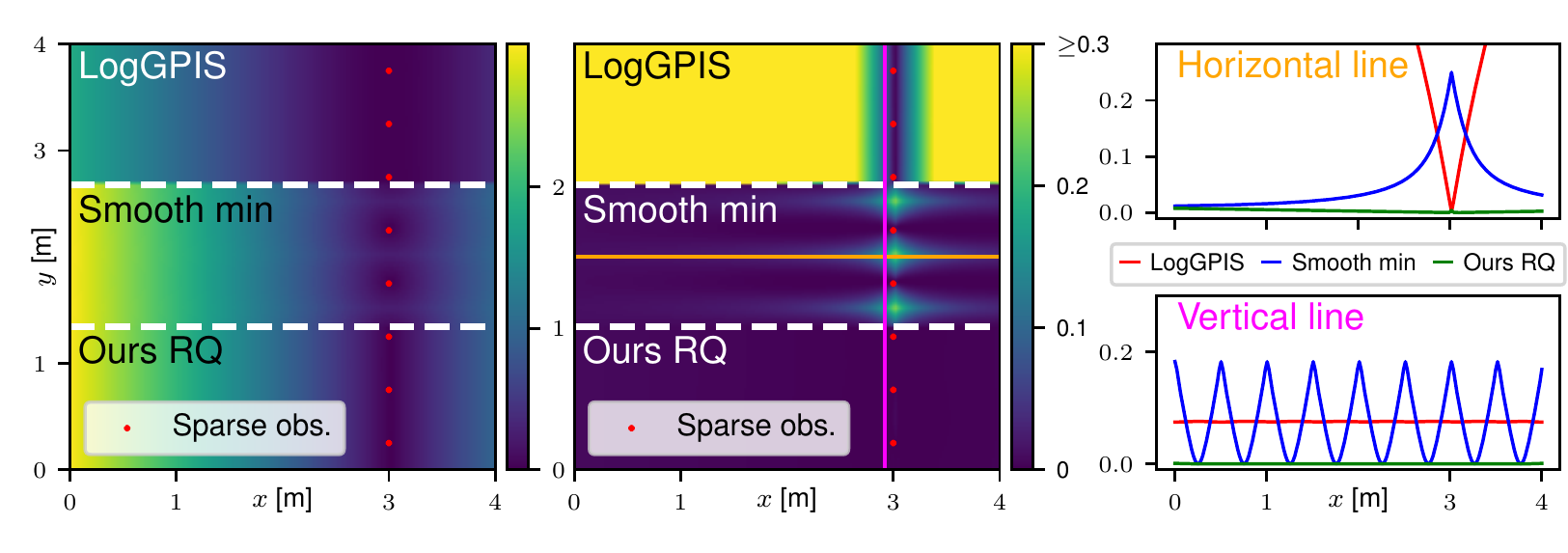}};

        \node[align = center, below=\legenddist of img,inner sep=0,outer sep=0, text width = 2.8cm, minimum width=3cm, xshift = -2.9cm, execute at begin node=\setlength{\baselineskip}{1.5ex}] (a) {\legendsize (a) Distance field estimates [m]};
        \node[align = center, right=\horispace of a,inner sep=0,outer sep=0, text width = 2.8cm, minimum width=3cm, execute at begin node=\setlength{\baselineskip}{1.5ex}] (b) {\legendsize (b) Distance field error [m]};
        \node[align = center, right=\horispace of b,inner sep=0,outer sep=0, text width = 2.8cm, minimum width=3cm, execute at begin node=\setlength{\baselineskip}{1.5ex}] (c) {\legendsize (c) Error along lines in (b)};
    \end{tikzpicture}
    \vspace{-0.6cm}
    \caption{Comparison of the interpolation abilities of different distance fields with sparse observations. Ours and LogGPIS use $\lengthscale = 0.75\units{m}$.}
    \label{fig:results_sparse}
\end{figure}
In this subsection, we benchmark the proposed method against LogGPIS \cite{Wu2021} and the simple smooth minimum function presented in Section~\ref{sec:smooth_min} over 100 simulated 2D environments.
Each environment is generated using the sum of sine functions with both random amplitudes and frequencies.
The typical gap between points is around $0.04\units{m}$.
To evaluate the different methods, we query 40k distance measurements per environment following a regular grid and compute the \ac{rmse} against the ground truth\footnote{The ground truth is generated by creating noiseless measurements of the surface at high density ($\approx 1\units{\mu m}$ between points) before finding the minimum distance between a query point and the dense surface points.}.
Table~\ref{tab:rmse_comparison} reports the average \ac{rmse} for each method considering separately close-range (under $0.05\units{m}$) and far-range (above $0.05\units{m}$) from the surface.
The last column corresponds to the success rate of each method (failure cases are due to numerical accuracy far from the surface).
Our hyperparameters have been tuned as presented in Section~\ref{sec:method}.
The same lengthscale has been used for LogGPIS to allow for the interpolation between the observations.
The parameter $\alpha$ of the \ac{rq} kernel and $\lambda$ of the smooth minimum have been selected with high values for accuracy but low enough to keep clear from machine precision at ranges considered in this experiment ($\alpha$=100,  $\lambda$ = -50).

Overall, our method consistently outperforms the other methods at close range.
Especially, one can see the high error of the smooth minimum corresponding to its inability to interpolate between the observations as illustrated in Fig.~\ref{fig:teaser} and~\ref{fig:results_sparse}.
These figures also illustrate the ability of the proposed method to accurately model the distance field despite noisy or sparse observations.
Please note that the value of the lengthscale $\lengthscale$ is changed to adapt to the sparsity of the data without hindering the performances away from the surface, unlike LogGPIS.

\begin{figure}
    \centering
    \includegraphics[width=\columnwidth, trim= 0.5cm 0cm 0.5cm 0cm]{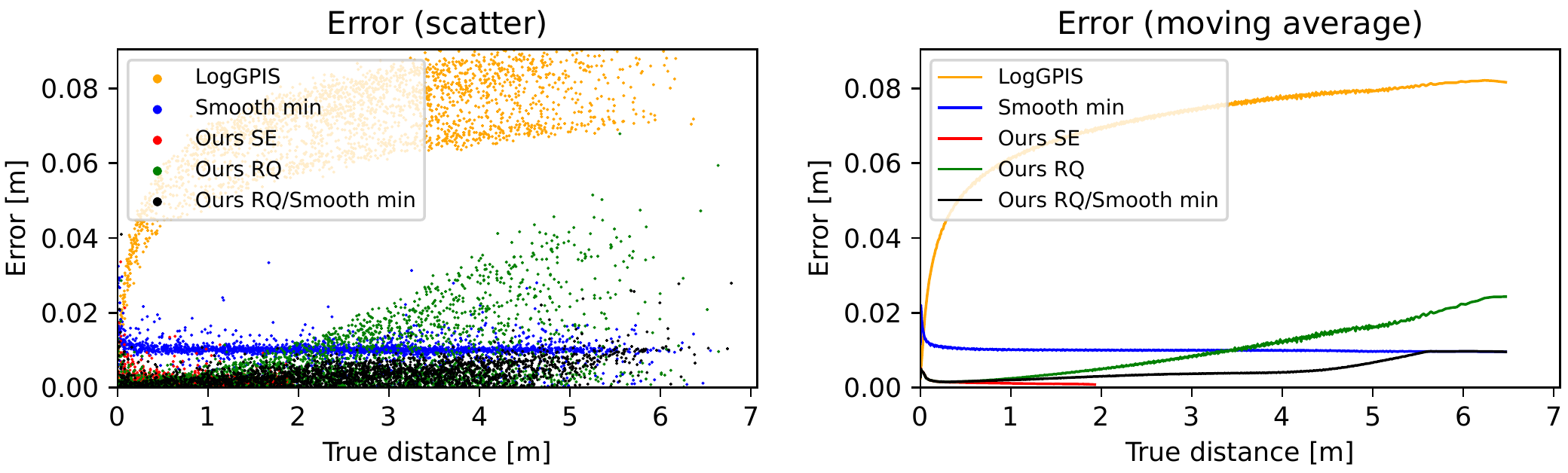}
    \vspace{-0.6cm}
    \caption{Comparison of the error distribution between LogGPIS \cite{Wu2021}, the smooth minimum function, and the proposed method with the RQ and SE kernels (2500 random samples among 100 simulated environments).}
    \label{fig:error_distribution}
\end{figure}

According to Table~\ref{tab:rmse_comparison}, the accuracy of the smooth minimum is on par with the proposed method when querying away from the surface.
In Fig.~\ref{fig:error_distribution} we display the absolute error of 2500 inferences randomly selected among the query points aforementioned.
Smoothing the plot with a moving average one can see that the error of LogGPIS follows a $\log(1+x)$ curve as predicted in Section~\ref{sec:background_loggpis}.
Fig.~\ref{fig:error_distribution} also introduces a merged approach combining the proposed method and the smooth minimum function leveraging the best of both methods: trusting the \ac{gp} reverting approach close to the surface and the smooth minimum further away with a smooth transition in between (progressive weighted average).
The overall accuracy of this fusion is better than each individual method as the smooth minimum tends to overestimate the distance due to its incapability to interpolate the surface, while our method can underestimate the field with potential ``over-interpolation" between distinct surfaces.

\subsection{Echolocation}
\label{subsec:Echolocation}
\begin{figure}
    \centering
    \def\scale{3.5cm}
    \def\hdist{0.1cm}
    \def\vdist{-0.05cm}
    \def\legendsize{\scriptsize}
    \begin{tikzpicture}
        \node[inner sep=0,outer sep=0] (traj) {\includegraphics[clip, trim=0cm 0.0cm 0cm 0.0cm, height=\scale]{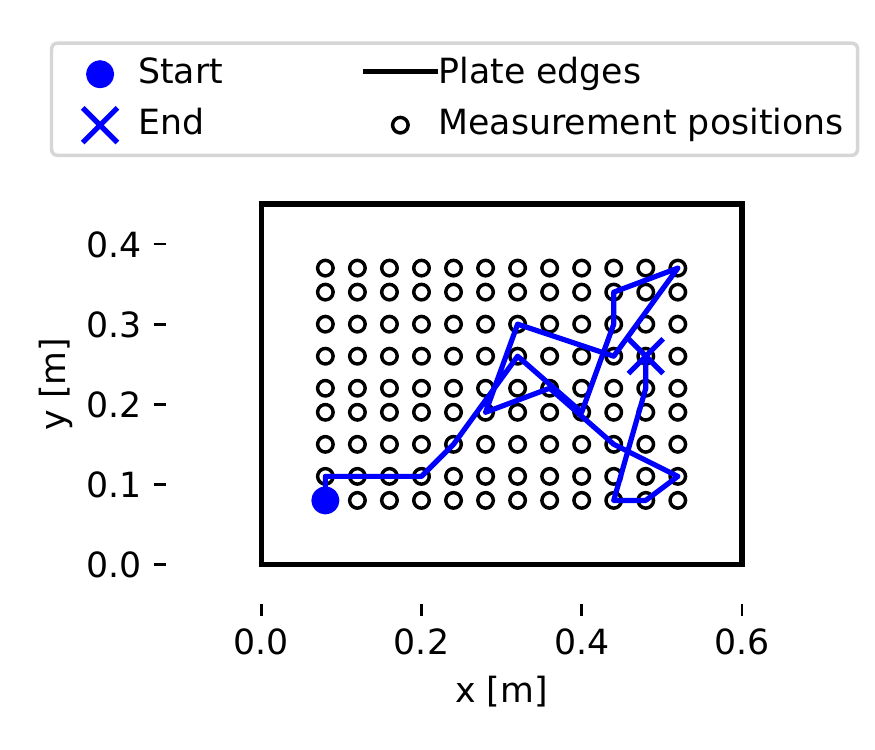}};
        \node[right=\hdist of traj,inner sep=0,outer sep=0] (plot) {\includegraphics[clip, trim=0cm 0.0cm 0cm 0.0cm, height=\scale]{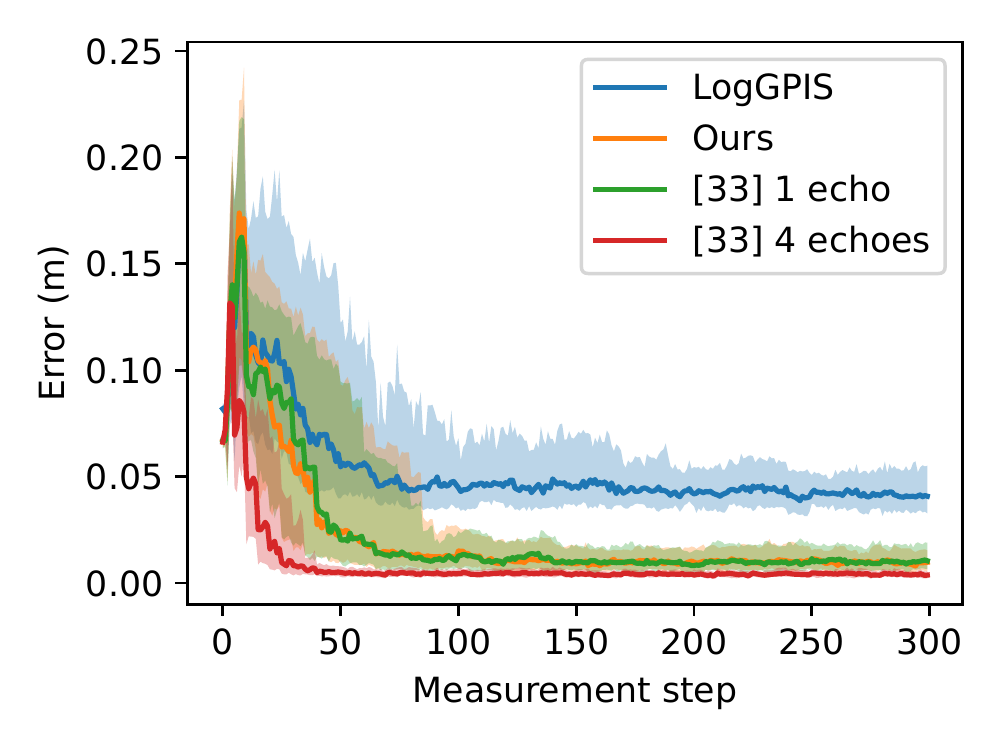}};
        \node[below=\vdist of traj,inner sep=0,outer sep=0] {\legendsize (a) 30-step trajectory example};
        \node[below=\vdist of plot,inner sep=0,outer sep=0] {\legendsize (b) Localisation error};
    \end{tikzpicture}
    \vspace{-0.6cm}
    \caption{Real-world localisation experiments. (a) illustrates the trajectory generation from a fixed grid of UGW measurements. (b) shows the median error obtained with several methods over 100 trajectories (the transparent areas correspond to the spread of the error between the $25^{th}$ and $75^{th}$ percentiles).} 
    \label{fig:localisation}
\end{figure}

In this subsection, we evaluate the proposed echolocation algorithm based on our \ac{gp} distance field (with \ac{rq} kernel and $\alpha=100$).
We have collected \ac{ugw} measurements in an aluminium metal panel with dimensions 600x450x6mm. 
The experimental setup consists of an emitter/receiver pair of nearly collocated transducers (as shown in Fig.~\ref{fig:teaser}) placed by hand on the vertices of a 9$\times$12 regular grid whose positions are carefully recorded.
For every position, the \ac{ugw} is generated using a two-tone burst sinusoidal wave at 100 kHz (excitation signal $s(t)$) while the receiver collects the ultrasonic response.
In total, 108 measurements were acquired.
From these grid-based measurements, we generated 100 trajectories of 300 steps by randomly navigation through the grid.
Fig.~\ref{fig:localisation}(a) illustrates 30 steps of such a trajectory.

We benchmark the localisation accuracy of the particle filter with LogGPIS, the particle filter with our novel reverting-function distance field, and the method from \cite{ouabi2021monte}.
In \cite{ouabi2021monte}, the environment is modelled with a set of geometric primitives (4 lines for our rectangular plate) and a particle filter is also used for localisation.
The parametric map approach allows for the computation of multiple echoes as part of the measurement model of each particle while our approach only considers the first echo.
For a more thorough comparison, we also implemented a version of \cite{ouabi2021monte} that considers solely the closest edge for each particle.
Note that it corresponds to the particle filter described in this work using the ground truth distance field of the rectangular panel.
We denote the two versions as \emph{\cite{ouabi2021monte} 1 echo} and \emph{\cite{ouabi2021monte} 4 echoes}.

Fig.~\ref{fig:localisation}(b) shows the median error as a function of the measurement step for the 100 trajectories.
The fact that \emph{\cite{ouabi2021monte} 1~echo} and \emph{Ours} result in very similar accuracy supports the evidence that the proposed representation approximates well the Euclidean distance field. 
It is not the case for LogGPIS which limits the overall system's performance in accordance with our simulated experiments in Section~\ref{sec:exp_dist_field}.
While providing satisfying levels of accuracy ($\approx 0.01\units{m}$), our method does not perform as well as \emph{\cite{ouabi2021monte} 4 echoes}.
This shows the importance of considering multiple echoes to improve the system's performance.
Future work will explore the integration of multiple echoes in our \ac{gp} representation.

\subsection{Mapping}

\begin{figure}
    \centering
    \includegraphics[clip, trim=1.5cm 0cm 1.5cm 0cm, width=0.99\columnwidth]{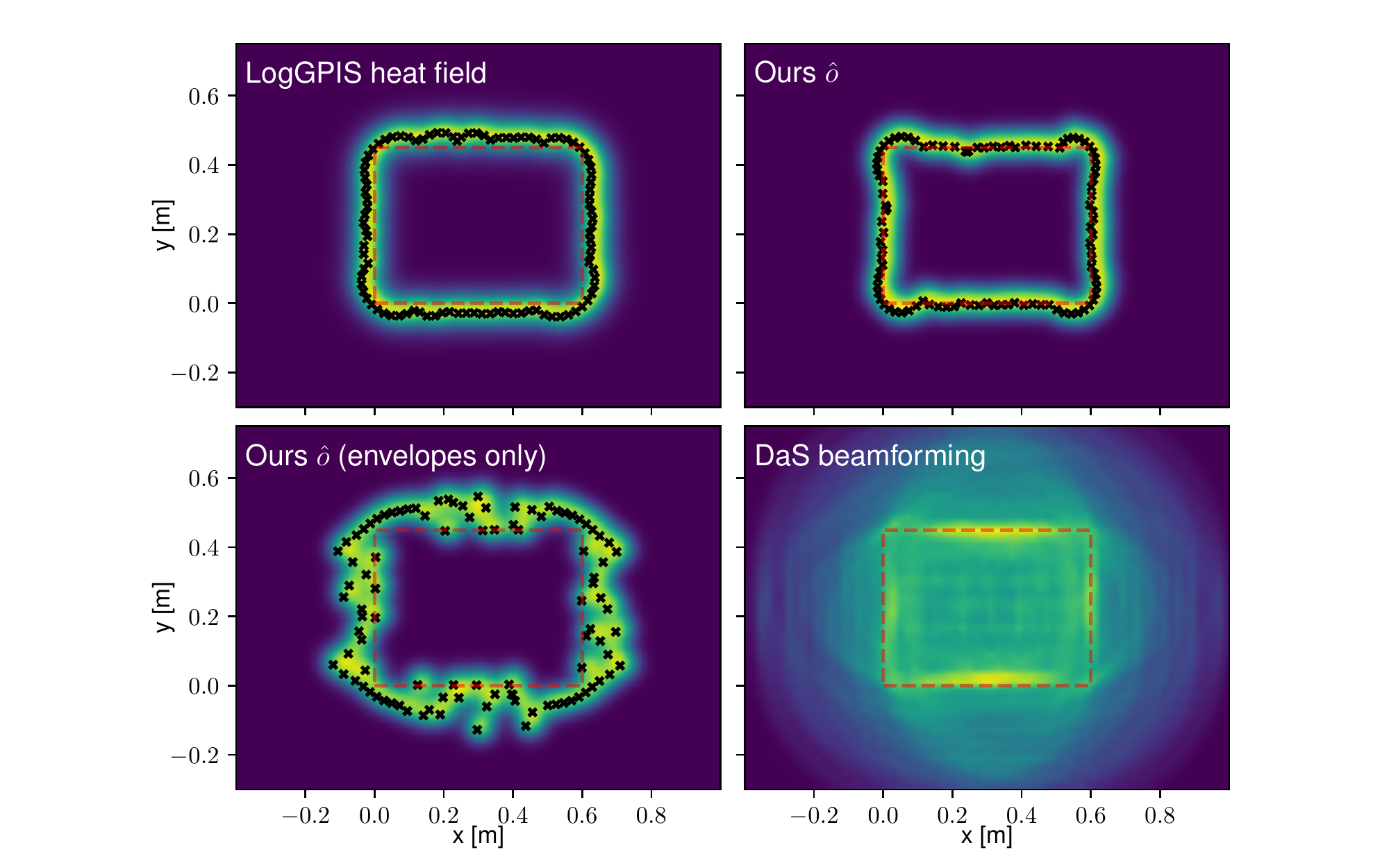}
    \vspace{-0.6cm}
    \caption{Latent field/reflector likelihood estimation for UGW-based mapping. The dashed line represents the ground truth of the metal panel's edges, and the crosses correspond to the optimised virtual observations $\abscissamat$.}
    \label{fig:mapping}
\end{figure}

To evaluate our \ac{ugw}-based mapping approach, we use the 108 measurements presented in the previous setup as input.
In Fig.~\ref{fig:mapping}, we show the different mapping results using LogGPIS, the proposed distance field (denoted \emph{ours}), and \ac{das} beamforming.
The former is a non-parametric approach for acoustic reflector mapping that assesses the likelihood of the presence of a reflector at any position $\abscissavec$ by summing the values of each envelope signal $e_i$ as $\mathcal{L}(\mathbf{x}) = \sum_{i=1}^N e_i(||\abscissavec_i - \abscissavec||)$.
As the \ac{das} approach does not provide a distance field, we chose to visualise the latent field of both LogGPIS and ours.
Note that the latent and distance fields are directly usable in downstream applications (localisation, path-planing, etc) whereas the \ac{das}-generated map is not.
Accordingly, the beamforming map needs additional processing to allow its use in autonomous systems.

To provide the reader with additional insight into our mapping pipeline, we run the mapping experiment without the first distance-based optimisation step \eqref{eq:dist_mapping} (denoted \emph{Ours (envelope-only)} in Fig.~\ref{fig:mapping}).
One can see that even if some estimated observations are present at the location of the plate's edges, the problem does not converge to the global minimum.
Converting the occupancy fields into distance fields and analysing the results inside the plate, we obtain distance \acp{rmse} of $0.021\units{m}$ for mapping using LogGPIS' distance field, $0.016\units{m}$ with \emph{Ours (envelope-only)}, and $0.006\units{m}$ with \emph{Ours}.
These results demonstrate the ability of our novel \ac{gp}-based distance to represent accurately the environment in a non-parametric manner while providing readily available information (occupancy and distance to obstacle) for later use of the mapping results.

\section{Conclusion}

In this paper, we presented a novel \ac{gp}-based distance field estimation method.
Using a point cloud as input, the proposed method first infers a latent field with \ac{gp} regression before applying a reverting function.
We empirically show that our approach significantly outperforms state-of-the-art representations with higher levels of accuracy allowing for novel applications which rely on precise Euclidean distance functions.
Accordingly, this work presents frameworks for echolocation and mapping using \acp{ugw}.
With real-world experiments, we demonstrated similar levels of localisation accuracy as those of parametric-based representations.
In the mapping context, our method does provide useful maps for various robotics applications (localisation, path planning, etc) while traditional \ac{das} maps need further processing.
Future work includes the development of a probabilistic shape reconstruction algorithm adapted to the proposed \ac{gp} latent field.
We will also explore the use of multi echoes in our \ac{gp} model to improve both \ac{ugw} echolocation and mapping in terms of robustness and accuracy.


\bibliographystyle{IEEEtran}
\bibliography{bibliography,sample}

\begin{acronym}[AAAAAAAAA]
    \acro{1d}[1D]{One-Dimensional}
    \acro{2d}[2D]{Two-Dimensional}
    \acro{3d}[3D]{Three-Dimensional}
    \acro{cas}[CAS]{Centre for Autonomous Systems}
    \acro{cpu}[CPU]{Central Processing Unit}
    \acro{das}[DaS]{Delay-and-Sum}
    \acro{dof}[DoF]{Degree-of-Freedom}
    \acro{dvs}[DVS]{Dynamic Vision Sensor}
    \acrodefplural{dvs}[DVS's]{Dynamic Vision Sensors}
    \acro{ekf}[EKF]{Extended Kalman filter}
	\acro{em}[EM]{Expectation-Maximization}
    \acro{esdf}[ESDF]{Euclidean Signed Distance Field}
    \acro{fov}[FoV]{Field-of-View}
    \acro{gnss}[GNSS]{Global Navigation Satellite System}
    \acrodefplural{gnss}[GNSS's]{Global Navigation Satellite Systems}
    \acro{gp}[GP]{Gaussian Process}
    \acrodefplural{gp}[GPs]{Gaussian Processes}
    \acro{gpm}[GPM]{Gaussian Preintegrated Measurement}
    \acro{ugpm}[UGPM]{Unified Gaussian Preintegrated Measurement}
    \acro{gps}[GPS]{Global Position System}
    \acrodefplural{gps}[GPS's]{Global Position Systems}
    \acro{gpis}[GPIS]{Gaussian Process Implicit Surface}
    \acro{gpu}[GPU]{Graphic Processing Unit}
    \acro{hdr}[HDR]{High Dynamic Range}
    \acro{icp}[ICP]{Iterative Closest Point}
    \acro{idol}[IDOL]{IMU-DVS Odometry using Lines}
    \acro{imu}[IMU]{Inertial Measurement Unit}
    \acro{in2laama}[IN2LAAMA]{INertial Lidar Localisation Autocalibration And MApping}
    \acro{kf}[KF]{Kalman Filter}
    \acro{kl}[KL]{Kullback¿Leibler}
    \acro{klt}[KLT]{Kanade–Lucas–Tomasi}
    \acro{lidar}[LiDAR]{Light Detection And Ranging Sensor}
    \acro{lpm}[LPM]{Linear Preintegrated Measurement}
    \acro{map}[MAP]{Maximum A Posteriori}
    \acro{mle}[MLE]{Maximum Likelihood Estimation}
    \acro{ndt}[NDT]{Normal Distribution Transform}
    \acro{pm}[PM]{Preintegrated Measurement}
    \acro{rrbt}[RRBT]{Rapidly Exploring Random Belief Trees}
    \acro{rgb}[RGB]{Red-Green-Blue}
    \acro{rgbd}[RGBD]{Red-Green-Blue-Depth}
    \acro{rms}[RMS]{Root Mean Squared}
    \acro{rmse}[RMSE]{Root Mean Squared Error}
    \acro{rq}[RQ]{Rational Quadratic}
    \acro{sde}[SDE]{Stochastic Differential Equation}
    \acro{SE3}[SE(3)]{Special Euclidean group in three dimensions}
    \acro{SE2}[SE(2)]{Special Euclidean group in 2D}
    \acro{se}[SE]{Square Exponential}
    \acro{slam}[SLAM]{Simultaneous Localisation And Mapping}
    \acro{so3}[$\mathfrak{so}$(3)]{Lie algebra of special orthonormal group in three dimensions}
    \acro{SO3}[SO(3)]{Special Orthonormal rotation group in three dimensions}
    \acro{tsdf}[TSDF]{Truncated Signed Distance Function}
    \acro{ugw}[UGW]{Ultrasonic Guided Wave}
    \acro{upm}[UPM]{Upsampled-Preintegrated-Measurement}
    \acro{uts}[UTS]{University of Technology, Sydney}
    \acro{vi}[VI]{Visual-Inertial}
    \acro{vio}[VIO]{Visual-Inertial Odometry}
    \acro{vo}[VO]{Visual Odometry}
 
\end{acronym}

\end{document}